\documentclass[letterpaper,journal,twoside]{IEEEtran}
\usepackage{amsmath,amsfonts}
\usepackage{algorithmic}
\usepackage{algorithm}
\usepackage{array}
\usepackage[caption=false,font=normalsize,labelfont=sf,textfont=sf]{subfig}
\usepackage{textcomp}
\usepackage{stfloats}
\usepackage{url}
\usepackage{verbatim}
\usepackage{graphicx}
\usepackage{cite}
\usepackage{amssymb}  
\usepackage{makecell}
\usepackage[table]{xcolor}
\usepackage{bbding}
\usepackage{booktabs}
\usepackage{hyperref}
\usepackage{threeparttable}
\usepackage{cleveref}
\usepackage{balance}
\usepackage{multirow}
\usepackage[normalem]{ulem}

\crefname{figure}{fig.}{figures}
\Crefname{figure}{Fig.}{Figures}

\begin{document}

\title{MVOFormer: Flow-Semantic Transformer for Robust Monocular Visual Odometry}

\markboth{Li \MakeLowercase{\textit{et al.}}: MVOFormer: Flow-Semantic Transformer for Robust Monocular Visual Odometry}%
{IEEE ROBOTICS AND AUTOMATION LETTERS. PREPRINT VERSION. ACCEPTED JUNE, 2026}
\makeatletter
\def\MVOFormerRALHeader{IEEE ROBOTICS AND AUTOMATION LETTERS. PREPRINT VERSION. ACCEPTED JUNE, 2026}
\def\MVOFormerAuthorHeader{Li \MakeLowercase{\textit{et al.}}: MVOFormer: Flow-Semantic Transformer for Robust Monocular Visual Odometry}
\def\ps@headings{%
\def\@oddhead{\hbox{}\@IEEEheaderstyle\MVOFormerAuthorHeader\hfil\thepage}\relax
\def\@evenhead{\@IEEEheaderstyle\thepage\hfil\MVOFormerRALHeader\hbox{}}\relax
\let\@oddfoot\@empty
\let\@evenfoot\@empty}
\pagestyle{headings}
\def\ps@IEEEtitlepagestyle{%
\def\@oddhead{\hbox{}\@IEEEheaderstyle\MVOFormerRALHeader\hfil\thepage}\relax
\def\@evenhead{\@IEEEheaderstyle\thepage\hfil\MVOFormerRALHeader\hbox{}}\relax
\let\@oddfoot\@empty
\let\@evenfoot\@empty}
\makeatother

\author{
	Jituo~Li\textsuperscript{1,*,\Envelope}, 
	Shunwang~Sun\textsuperscript{1,*}, 
    Jialu~Zhang\textsuperscript{1}, Xinqi~Liu\textsuperscript{2,\Envelope}, 
    Jinyao Hu\textsuperscript{1}, Zhicheng Lu\textsuperscript{3}, Sajad Saeedi\textsuperscript{4}, Guodong~Lu\textsuperscript{1}
    
	\thanks{
    Manuscript received: February 20, 2026; Revised: May 11, 2026; Accepted: June 12, 2026. This paper was recommended for publication by Editor Ayoung Kim upon evaluation of the Associate Editor and Reviewers' comments.\par
    This work was supported in part by the ``Pioneer\& Leading Goose" R\&D Program of Zhejiang Province, China under Grant 2023C01067, and in part by the State Key Laboratory of Digital-Intelligent Modeling and Simulation, and Funding of Zhejiang University Robotics Institute. \textsuperscript{*}Equal contribution. \textsuperscript{\Envelope}Corresponding authors: Jituo Li and Xinqi Liu. 

    \textsuperscript{1}Jituo Li, Shunwang Sun, Jialu Zhang, Jinyao Hu and Guodong Lu are with State Key Laboratory of Fluid Power and Mechatronic Systems, Zhejiang University and with Zhejiang Key Laboratory of Industrial Big Data and Robot Intelligent Systems. They are also with the School of Mechanical Engineering, Zhejiang University, Hangzhou 310027, China, and with Robotics Institute, Zhejiang University, Hangzhou 310027, China (e-mail: jituo\_li@zju.edu.cn).
    
    \textsuperscript{2}Xinqi Liu is with the School of Artificial Intelligence and Robotics, Hunan University, Changsha, 410012, China (e-mail: liuxinqi@hnu.edu.cn).
    
    \textsuperscript{3}Zhicheng Lu is with the Rural Health Research Institute, Charles Sturt University, Sydney, Australia.

    \textsuperscript{4}Sajad Saeedi is with University College London, U.K.

    Digital Object Identifier (DOI): see top of this page.
	}
}
\maketitle

\begin{abstract}

Monocular visual odometry (MVO) is foundational to autonomous navigation and robotic localization. However, existing learning-based MVO approaches often struggle with either a lack of interpretable, complementary features or overly complex multi-stage architectures. These limitations inherently restrict their robustness and cross-domain generalization. In this work, we propose MVOFormer, a novel transformer framework for robust monocular visual odometry. Our architecture features a Flow-Semantic Dual Branch Encoder that synergizes dense geometric motion cues with object-centric semantic priors, explicitly distinguishing static structures from dynamic distractors. These representations are then fused by an Iterative Multimodal Decoder, enabling coarse-to-fine pose refinement while dynamically suppressing attention on unreliable regions. Extensive evaluations demonstrate that, without any target-domain fine-tuning, MVOFormer achieves superior zero-shot generalization and robustness, significantly outperforming prior learning-based frame-to-frame methods across diverse benchmarks including TartanAir, KITTI, TUM-RGBD, and ETH3D-SLAM. Code and model checkpoints will be made available at https://github.com/Sun-Shun/MVOFormer.

\end{abstract}


\section{Introduction}
\label{sec1}

\begin{figure*}[htbp]
\centering
\includegraphics[scale=0.375]{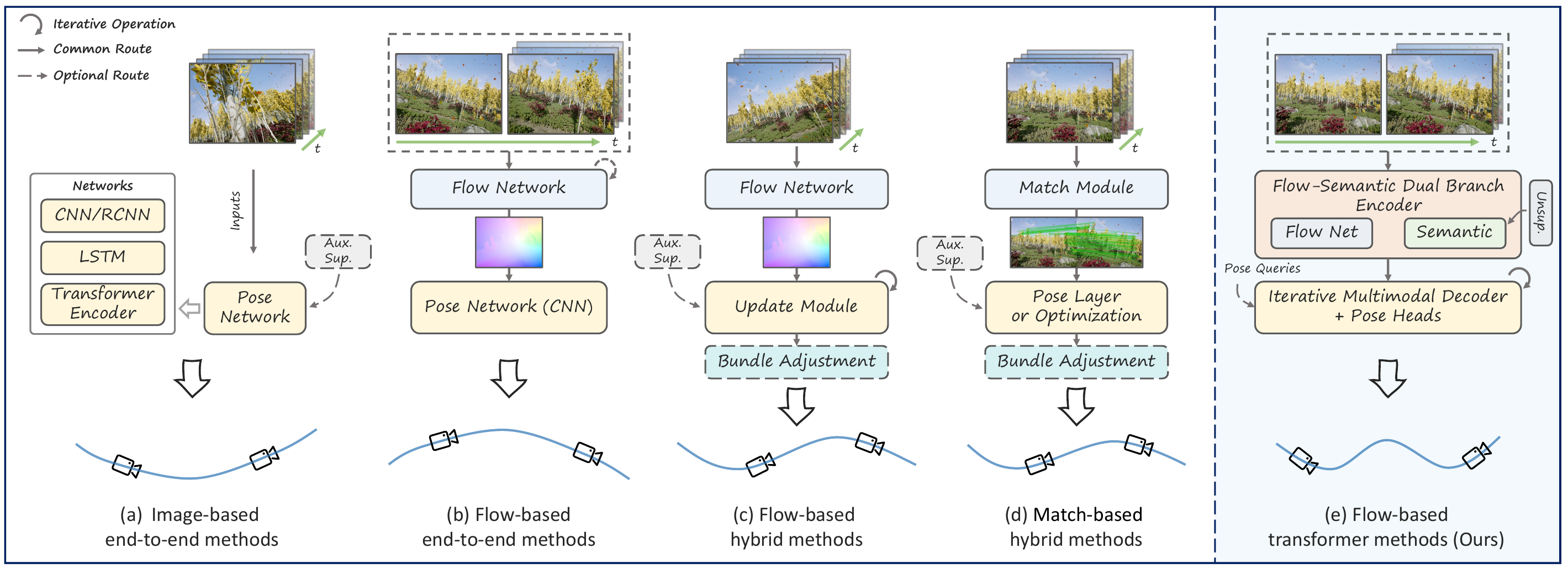}
\caption{\textbf{Comparison of contemporary learning-based MVO methods.}
\textcolor[rgb]{1., .824, .325}{\textbf{Yellow boxes}} indicate learnable pose network modules.
``Aux. Sup.'' denotes methods requiring auxiliary supervision (e.g., semantics, depth), while ``Unsup.'' represents unsupervised approaches.
\textbf{Dashed lines} indicate components utilized by only a subset of methods within the category.
}\label{fig1}
\end{figure*}

\IEEEPARstart{M}{onocular} visual odometry estimates camera ego-motion from a single image stream and forms a core component of visual SLAM, autonomous driving, AR/VR, and robotic navigation. Classical methods~\cite{campos2021orb,forster2014svo,engel2017direct} rely on hand-crafted multi-stage pipelines (feature extraction, tracking, outlier rejection, and bundle adjustment), which are brittle under illumination changes, low texture, or dynamic objects and require careful parameter tuning.

Deep learning offers superior robustness and simplicity, and current deep VO methods can be broadly divided into direct pose regression and hybrid paradigms (\Cref{fig1}). Early works employed recurrent convolutional networks~\cite{wang2017deepvo} to directly model camera motion dynamics from image sequences. The advent of transformers has further inspired a series of VO architectures~\cite{wu2024swformer,zhao2022transformer,chiu2023vitvo}, which benefit from powerful global context modeling capabilities. Despite these advances, transformer-based methods continue to exhibit limited zero-shot capability when deployed outside their training distributions. A key limitation lies in their reliance on a single global class token to regress poses from noisy image pairs, which leads to a lack of interpretable representations. To alleviate domain shift across diverse environments, incorporating optical flow has proven highly effective~\cite{wang2021tartanvo}. As shown in \Cref{fig1}~(b), TartanVO~\cite{wang2021tartanvo} and DytanVO~\cite{shen2023dytanvo} introduced auxiliary optical flow prediction as a self-supervised signal, significantly enhancing cross-dataset generalization. 

Despite its advantages, optical flow inherently captures only 2D pixel-level motion and lacks object-level semantic context, which can lead to ambiguous or failed motion estimation in textureless or homogeneous regions where pixel appearance provides insufficient discriminative cues.

In contrast, recent hybrid methods mitigate these limitations by explicitly injecting semantic or geometric priors. For instance, flow-based hybrid methods~\cite{teed2021droid,teed2023deep} (\Cref{fig1}~(c)) iteratively refine dense correspondences through recurrent update operators. Similarly, PVO~\cite{ye2023pvo} integrates semantic segmentation with visual odometry, yielding strong performance. However, it relies on multi-model training, requires extra semantic labels, and employs three heavily intertwined modules, which significantly complicates both training and inference. More recently, matching-based hybrid methods~\cite{wang2025mambavo,qiu2025mac} (\Cref{fig1}~(d)) exploit semantic features~\cite{oquab2023dinov2} or depth to achieve robust pixel correspondence. Despite their effectiveness, these hybrid approaches invariably rely on multi-stage or multi-model pipelines, require meticulous per-component hyperparameter tuning, and entail substantially more intricate training and inference procedures than direct pose regression methods.

To bridge this gap, we propose \textbf{MVOFormer}, a novel transformer-based framework (\Cref{fig1}(e)) that preserves architectural simplicity while injecting explicit geometric and semantic priors for more robust motion estimation. As illustrated in \Cref{fig2}, we introduce a \textbf{Flow-Semantic Dual Branch Encoder} to jointly extract dense optical flow and object-centric semantics. An \textbf{Iterative Multimodal Decoder} then fuses these complementary cues, progressively suppressing dynamic objects and refining 6-DoF poses in a coarse-to-fine manner. Crucially, MVOFormer directly estimates poses from raw image pairs in a single forward pass, achieving highly competitive results without target-domain fine-tuning.

Our primary contributions are summarized as follows:

\begin{itemize}

\item We propose MVOFormer, a novel transformer-based MVO framework without auxiliary semantic segmentation or depth supervision.
By seamlessly integrating two complementary geometric cues into a unified architecture, MVOFormer delivers superior accuracy and remarkable zero-shot capability without multi-model training.

\item We introduce a Flow-Semantic Dual Branch Encoder that jointly captures 2D motion correspondences and object-level semantic priors, thereby enriching object perception and distinguishing dynamic regions from static ones.

\item We design an Iterative Multimodal Decoder that seamlessly fuses these features for coarse-to-fine pose refinement while actively suppressing dynamic distractors.

\item Extensive experiments on TartanAir~\cite{wang2020tartanair}, KITTI~\cite{geiger2012we}, TUM-RGBD~\cite{sturm2012benchmark}, and ETH3D-SLAM~\cite{schops2019bad} demonstrate that our framework outperforms existing learning-based frame-to-frame methods and exhibits remarkable robustness in dynamic scenes.

\end{itemize}

\section{Related Work}
\label{sec2}
\subsection{Visual Odometry}

\textbf{Classical Methods.} Classical VO methods fall into feature-based~\cite{campos2021orb}, direct~\cite{engel2017direct}, and semi-direct~\cite{forster2014svo} categories. However, these hand-crafted systems are fundamentally brittle: feature-based models drift in low-texture or dynamic scenes, direct methods fail under illumination changes, and semi-direct approaches require fragile parameter tuning. These limitations heavily motivate the shift toward learning-based VO.

\textbf{Hybrid Methods.} These systems~\cite{teed2021droid,teed2023deep} combine learned correspondence matching with classical optimization (e.g., bundle adjustment). While demonstrating impressive accuracy, their multi-stage nature introduces architectural complexity, which sacrifices the elegance of direct inference and remains sensitive to erroneous dense correspondences.

\textbf{Direct Pose Regression Methods.} These models directly regress poses from image sequences. Early CNN-based works~\cite{wang2017deepvo,konda2015learning} suffered from poor zero-shot capability. TartanVO~\cite{wang2021tartanvo} alleviated this by introducing auxiliary flow supervision to enhance generalization. Recent transformer-based models~\cite{chiu2023vitvo,zhao2022transformer,wu2024swformer} exploit self-attention for spatio-temporal modeling and object suppression, yet most are trained and evaluated in-domain and show limited generalization. In contrast, MVOFormer provides a unified solution, seamlessly integrating geometric flow with semantic priors to deliver robust zero-shot accuracy while retaining inference simplicity.

\begin{figure*}[htbp]
\centering
\includegraphics[scale=0.38]{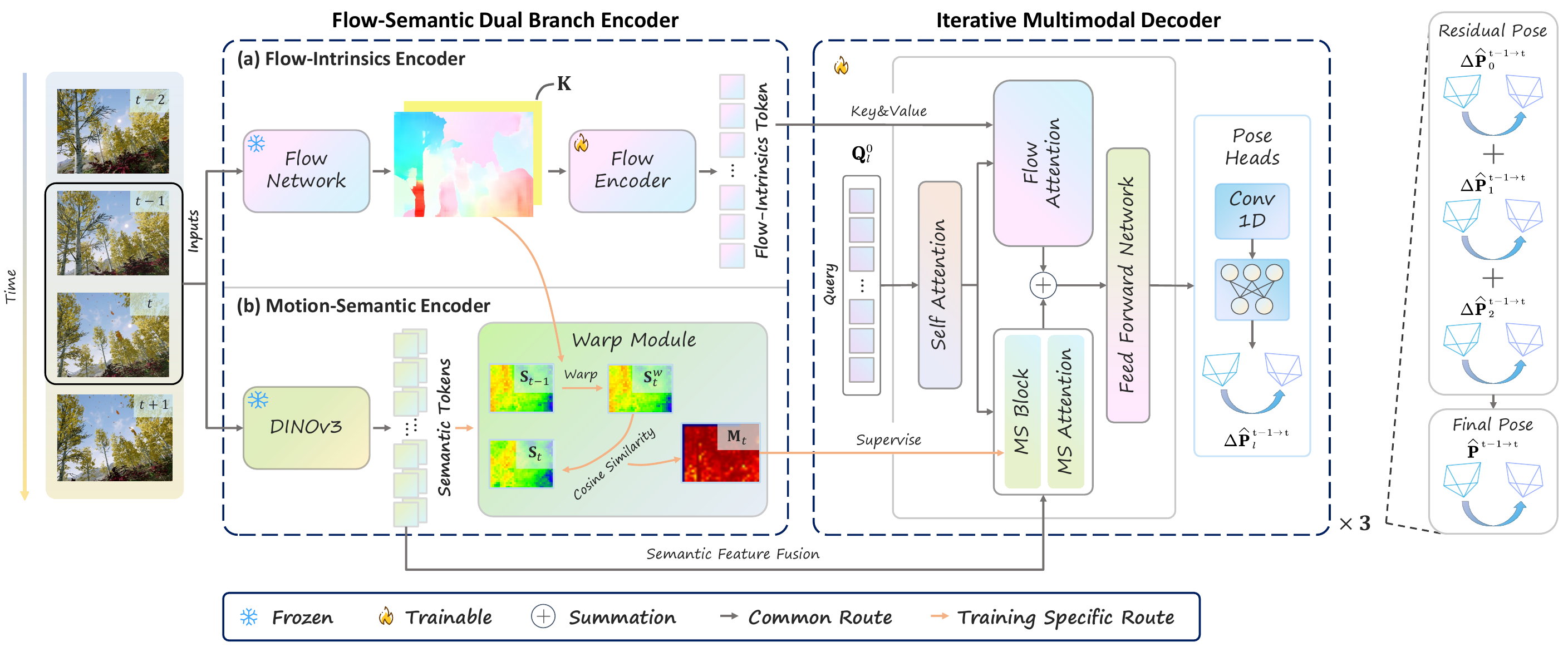}
\caption{Overview of the MVOFormer architecture. It consists of a Flow-Semantic Dual Branch Encoder comprising the Flow-Intrinsics Encoder and the Motion-Semantic Encoder, an Iterative Multimodal Decoder, and the final Pose Heads. Note that the training-specific route is deactivated during inference, introducing no computational overhead during deployment. During training, the Warp Module is activated to generate the supervisory signal for the Uncertainty Module within the MS Block.
}\label{fig2}
\end{figure*}

\subsection{Multi-Task and Multimodal Visual Odometry}
Recent works combine VO with dense predictions (semantics, depth, flow) for richer supervision. Specifically, systems jointly predict panoptic/instance segmentation~\cite{ye2023pvo,xie2024instancevo}, couple VO with dense reconstruction~\cite{naumann2024nerf}, or exploit temporal consistency and pretrained priors~\cite{shen2023dytanvo,zhang2025leveraging,wei2024bev,wang2025mambavo} to effectively handle dynamic objects.

Despite their effectiveness, these hybrid methods often rely on multi-stage pipelines, require explicit semantic labels, or introduce architectural redundancy. In contrast, MVOFormer adopts a compact transformer architecture with direct single-pass pose regression. By seamlessly fusing geometric flow and semantic priors, it delivers robust zero-shot capability without relying on auxiliary supervision or complex optimization.

\section{Methods}
\label{sec3}
\subsection{Overview}
\label{subsec3.1}
As illustrated in \Cref{fig2}, given a monocular sequence $\{I_t\}_{t=0}^T$, MVOFormer directly estimates 6-DoF relative poses $\mathbf{P}^{t-1\to t}={\rm{F_{VO}}}(I_{t-1}, I_t)$ in a single forward pass. To overcome the limited interpretability and poor zero-shot generalization of direct RGB regression, our framework comprises two core transformer modules. 

First, the \textbf{Flow-Semantic Dual Branch Encoder} (\Cref{subsec3.2}) extracts complementary features: a Flow-Intrinsics branch (Sea-RAFT~\cite{wang2024sea}) captures dense geometric motion, while a Motion-Semantic branch (frozen DINOv3~\cite{simeoni2025dinov3}) provides object-centric tokens. A Warp Module then fuses these streams into a static similarity map to detect dynamic regions. 
Second, the \textbf{Iterative Multimodal Decoder} (\Cref{subsec3.3}) fuses these representations to iteratively refine pose queries across three layers, actively utilizing uncertainty estimates to downweight dynamic distractors. Finally, Pose Heads (\Cref{subsec3.4}) regress the relative residuals $\Delta\mathbf{P}^{t-1\to t}$ to yield the final pose. Training objectives are detailed in \Cref{subsec3.5}.

\subsection{Flow-Semantic Dual Branch Encoder}
\label{subsec3.2}
\begin{equation} \label{eq1}
\mathbf{F}_{t-1 \to t} = \textbf{FlowNet}(I_{t-1}, I_t)
\end{equation}

We convert the camera intrinsics $\{f_x,f_y,o_x,o_y\}$ into a full-resolution map $\mathbf{K}\in\mathbb{R}^{H\times W\times 2}$ via bilinear indexing. Following TartanVO~\cite{wang2021tartanvo}, random cropping is applied to both the optical flow $\mathbf{F}_{t-1\to t}\in\mathbb{R}^{H\times W\times 2}$ and $\mathbf{K}$, effectively simulating diverse camera parameters and boosting generalization.

The extracted flow and camera intrinsics are channel-wise concatenated, downsampled to 1/4 resolution, and subsequently processed by a trainable ResNet-50 backbone. To conserve memory while focusing on sparse informative regions~\cite{chiu2023vitvo}, the resulting features are augmented with positional embeddings and refined through three multi-scale deformable attention blocks~\cite{zhu2020deformable}. This branch ultimately yields Flow-Intrinsics Tokens $\mathbf{T}_{\text{flow}}^{t-1\to t}$, equipping the downstream decoder with geometrically precise motion cues.

\begin{equation} \label{eqflowencoder}
\mathbf{T}_{\text{flow}}^{t-1\to t} = \textbf{FlowEncoder}(\mathbf{F}_{t-1\to t}, \mathbf{K})
\end{equation}

\textbf{Motion-Semantic Encoder.} As depicted in \Cref{fig2} (b), each frame $I_{t-1}$ and $I_t$ is independently processed by the frozen DINOv3 backbone, producing dense semantic tokens $\mathbf{T}_\text{{MS}}=\{\mathbf{S}_{t-1}, \mathbf{S}_t\}, \mathbf{S}_t \in \mathbb{R}^{\frac{H}{n} \times \frac{W}{n} \times C_\text{DINO}} (C_\text{DINO}=384)$. DINOv3's features are highly robust and object-centric, providing strong high-level scene understanding for subsequent dynamic-region suppression and multimodal fusion.

To obtain an explicit motion mask without any labeled data, we warp the previous-frame features $\mathbf{S}_{t-1}$ toward the current frame using the optical flow $\mathbf{F}_{t-1 \to t}$ from Sea-RAFT:
\begin{equation} \label{eq3}
\mathbf{S}_t^w = \textbf{Warp}(\mathbf{S}_{t-1}, \mathbf{F}_{t-1 \to t})
\end{equation}
where $\mathbf{F}_{t-1 \to t}$ is first resized to match the spatial resolution of $\mathbf{S}_{t-1}$, before warping.

As shown in \Cref{fig2}, we compute per-pixel cosine similarity between $\mathbf{S}_t$ and their flow-warped counterparts:
\begin{equation} \label{eq4}
\mathbf{M}_t = \cos(\mathbf{S}_t, \mathbf{S}_t^w) = \frac{\mathbf{S}_t \odot \mathbf{S}_t^w}{\|\mathbf{S}_t\| \|\mathbf{S}_t^w\|} \in [-1, 1]
\end{equation}
where $\odot$ denotes element-wise multiplication. The similarity is computed only over valid warped pixels. As expected, $\mathbf{M}_t$ yields high values on static regions and low values on dynamic objects, thereby providing a robust, self-supervised confidence weight for our uncertainty-aware loss.

\begin{figure}[!t]
\centering
\includegraphics[width=\linewidth]{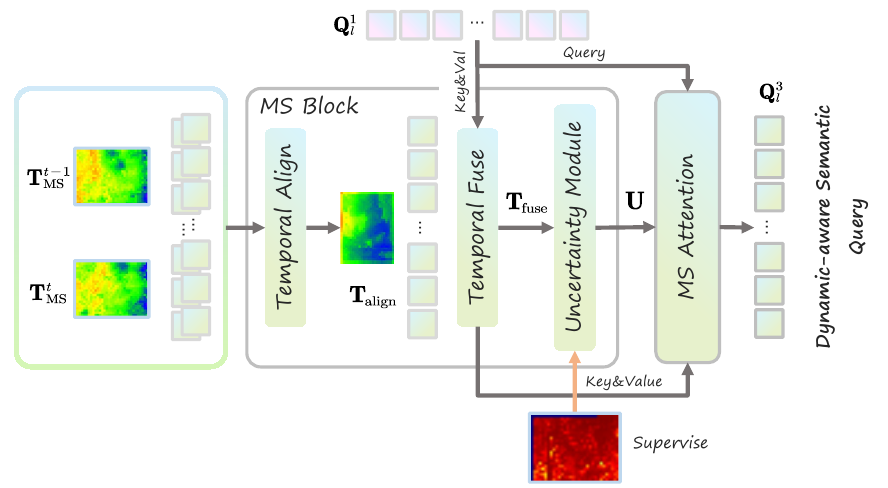}
\caption{Dynamic-aware processing in the Motion-Semantic (MS) branch. Temporal Align aligns per-frame DINOv3 tokens, Temporal Fuse aggregates them, and Uncertainty Module generates a self-supervised confidence map $\mathbf{W}_{\text{mask}}$.}\label{fig4}
\end{figure}

\subsection{Iterative Multimodal Decoder}
\label{subsec3.3}
The Iterative Multimodal Decoder (\Cref{fig2}) is designed to fuse flow-intrinsics tokens $\mathbf{T}_{\text{flow}}$ and motion-semantic tokens $[\mathbf{T}_{\text{MS}}^{t-1},\mathbf{T}_{\text{MS}}^t]$. It consists of $L=3$ stacked layers that progressively refine $N=100$ learnable pose queries $\mathbf{Q}_l^0 \in \mathbb{R}^{N \times C}$ ($C=256$) for coarse-to-fine pose estimation. Guided by the Warp Module's similarity maps during training, the decoder effectively suppresses attention on irrelevant dynamic regions. Specifically, each layer comprises the following sequentially executed components:

\textbf{(1) Self-Attention.} Standard multi-head self-attention with 8 heads and residual connection:
\begin{equation} \label{eq5}
\mathbf{Q}_l^1 = \text{MHSAttn}(\mathbf{Q}_l^0)
\end{equation}

\textbf{(2) Flow Attention.} We address the problem of attention overly focusing on a sparse set of pixels~\cite{chiu2023vitvo} by employing multi-scale deformable cross-attention. Specifically, decoder queries attend to the low-level Flow-Intrinsics Tokens:
\begin{equation} \label{eq6}
\mathbf{Q}_l^2 = \text{DeformCrossAttn}(\mathbf{Q}_l^1, \mathbf{T}_{\text{flow}}^{t-1\to t}) + \mathbf{Q}_l^1
\end{equation}

\textbf{(3) Motion-Semantic (MS) Block.} As shown in \Cref{fig4}, this block generates robust, dynamic-aware semantic tokens before the final cross-attention. It consists of three components.

\textbf{Temporal Align Module:} Inspired by PanoOcc~\cite{wang2024panoocc}, we replace their computationally intensive 3D voxel convolutions with efficient 2D variants to implicitly align spatio-temporal contexts. It takes the adjacent Semantic Tokens $[\mathbf{T}_{\text{MS}}^{t-1},\mathbf{T}_{\text{MS}}^t]$ as input and outputs the aligned temporal semantic features $\mathbf{T}_\text{{align}}\in \mathbb{R}^{\frac{H}{n} \times \frac{W}{n} \times C_\text{DINO}}$:
\begin{equation}
    \mathbf{T}_\text{{align}} = \text{Bottleneck2D}([\mathbf{T}_{\text{MS}}^{t-1} \parallel \mathbf{T}_{\text{MS}}^t]) 
\end{equation}
where $[\cdot \Vert \cdot]$ denotes channel-wise concatenation. $\text{Bottleneck}_{2D}$ is a lightweight residual 2D bottleneck block. It captures local temporal dynamics.

\textbf{Temporal Fuse Module:} A multi-head cross-attention layer models the interaction between the query and the aligned features, yielding the fused output $\mathbf{T}_\text{{fuse}}$:

\begin{equation} \label{eq7}
\mathbf{T}_\text{{fuse}} = \text{MHCAttn}(\mathbf{T}_\text{{align}}, \mathbf{Q}_l^1)
\end{equation}
where $\mathbf{T}_\text{{align}}$ as the query and $\mathbf{Q}_l^1$ as the key and value in this cross-attention. This assignment ensures that the temporal semantic feature $\mathbf{T}_\text{{align}}$ is dominant in the resulting tokens. 

\textbf{Uncertainty Module.} Inspired by the probabilistic formulation in~\cite{kulhanek2024wildgaussians}, this module predicts a continuous per-patch uncertainty map $\mathbf{U} \in \mathbb{R}^{\frac{H}{n} \times \frac{W}{n}}$. The map down-weights contributions from moving objects. The fused feature $\mathbf{T}_\text{fuse}$ is processed by spatial dropout, batch normalization, and a $1\times1$ convolution to produce a scalar map. We then apply a Softplus activation with a shifted prior for positivity and numerical stability:
\begin{equation}
    \mathbf{U} = \text{Softplus}\big(\text{Conv2D}_{1\times 1}(\mathbf{T}_\text{fuse}) + \ln(e-1)\big).
\end{equation}

The predicted uncertainty $\mathbf{U}$ generates the dynamic-aware modulation mask $\mathbf{W}_\text{mask}$. To prevent gradient explosion, we make the masking weights inversely proportional to squared uncertainty and clip them at 3:
\begin{equation}
    \mathbf{W}_\text{mask} = \min\left(\frac{1}{2 \max(0.1, \mathbf{U})^2}, \, 3\right).
\end{equation}

\textbf{(4) Motion-Semantic Attention (MS Attention).} Queries attend to the processed semantic tokens $\mathbf{T}_\text{{fuse}}$, modulated by the learned uncertainty mask:

\begin{equation} \label{eq8}
\begin{aligned}
\mathbf{Q}_l^3 &= \text{MHCAttn}(\mathbf{Q}_l^1, \mathbf{T}_\text{fuse}, \mathbf{W}_\text{mask}) \\
&= \text{Softmax}\left(\frac{QK^\top}{\sqrt{d}} + \log \mathbf{W}_{\text{mask}}\right) V
\end{aligned}
\end{equation}
where $\mathbf{T}_\text{{fuse}}$ is used as both key and value. The mask $\mathbf{W}_\text{mask}$ reweights the attention scores. This helps the network prioritize stable, semantically rich regions and suppress dynamic or unreliable distractors.

\textbf{(5) Feed-Forward Network (FFN).} The flow-refined query $\mathbf{Q}_l^2$ and the dynamic-aware semantic query $\mathbf{Q}_l^3$ are fused and passed to the FFN. The output is the pose query $\mathbf{Q}_{l}^{\text{pose}}$ for the current decoder block, which encodes both flow cues and motion-masked semantic information:
\begin{equation} \label{eq9}
\mathbf{Q}_{l}^\text{pose} = \text{FFN}(\mathbf{Q}_l^2 + \gamma \mathbf{Q}_l^3)
\end{equation}
where $\gamma$ is the weight assigned to the motion-semantic query, fixed at $\gamma = 0.8$.

\subsection{Pose Heads}
\label{subsec3.4}
To enable explicit coarse-to-fine refinement, we attach a shared lightweight pose regression head to the output queries of each decoder layer $l \in \{1, 2, 3\}$. Each pose head predicts a residual pose update $\Delta \hat{\mathbf{P}}_l^{t-1 \to t} \in \mathfrak{se}(3)$, which is parameterized as $\Delta \hat{\mathbf{P}}_l^{t-1 \to t} = (\Delta \hat{\mathbf{T}}_l^{t-1 \to t}, \Delta \hat{\mathbf{R}}_l^{t-1 \to t})$, where $\Delta \hat{\mathbf{T}}_l^{t-1 \to t} \in \mathbb{R}^3$ is the 3D translation and $\Delta \hat{\mathbf{R}}_l^{t-1 \to t} \in \mathfrak{so}(3)$ is the 3D rotation component:
\begin{equation} \label{eq10}
\Delta \hat{\mathbf{P}}_l^{t-1 \to t} = \mathbf{Pose}_{\theta_{l, k}}(\mathbf{Q}_{l}^\text{pose})
\end{equation}
where the pose head network $\mathbf{Pose}_{\theta_{l, k}}(\mathbf{Q}_{l}^\text{pose})$ is constructed using a 1D convolution followed by a 3-layer MLP as illustrated in \Cref{fig2}. For the two pose components, translation $\Delta \hat{\mathbf{T}}_l^{t-1 \to t}$ and rotation $\Delta \hat{\mathbf{R}}_l^{t-1 \to t}$, we employ two distinct prediction networks, indexed by $k \in \{1, 2\}$, each utilizing different parameters $\theta_{l, k}$.

The final pose $\hat{\mathbf{P}}^{t-1 \to t}$ is obtained by sequentially accumulating the predicted pose adjustments from all decoder layers, thereby realizing the coarse-to-fine iterative reasoning process:

\begin{equation} \label{eq11}
\hat{\mathbf{P}}^{t-1 \to t} = \sum_{l=0}^2 \Delta \hat{\mathbf{P}}_l^{t-1 \to t}
\end{equation}

This progressive residual formulation allows early layers to capture coarse motion primarily from flow tokens, while deeper layers incorporate finer semantic corrections and dynamic suppression, resulting in superior robustness. 

\begin{figure}[!t]
\centering
\includegraphics[scale=0.43]{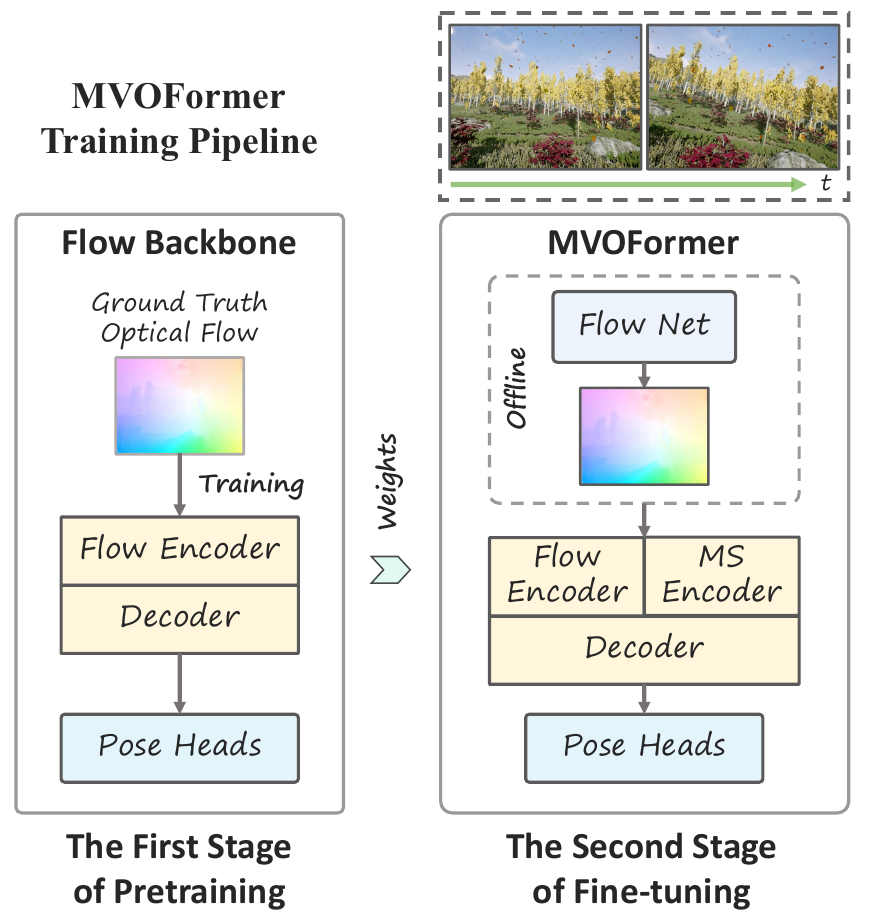}
\caption{Two-stage training pipeline of MVOFormer.}
\label{fig6}
\end{figure}

\subsection{Loss Functions}
\label{subsec3.5}
\textbf{Pose Loss.}
Following the learning-based MVO~\cite{wang2021tartanvo}, we supervise the relative pose at every decoder layer using a weighted combination of translation and rotation errors. For layer $l \in \{0,1,2\}$, the per-layer pose loss is defined as
\begin{align} \label{eq12}
\nonumber
\mathcal{L}^{\text{pose}}_l =& \|  \frac{\Delta \hat{\mathbf{T}}^{t-1 \to t}_l}{\max(\Vert \Delta \hat{\mathbf{T}}^{t-1 \to t}_l \Vert, \epsilon)} - \frac{\Delta \mathbf{T}^{t-1 \to t}_{\text{gt}}}{\max(\Vert \Delta \mathbf{T}^{t-1 \to t}_{\text{gt}} \Vert, \epsilon)} \|_1 \\
&+ \| \Delta \hat{\mathbf{R}}^{t-1 \to t}_l - \Delta \mathbf{R}^{t-1 \to t}_{\text{gt}} \|_1
\end{align}

To handle monocular scale ambiguity, we normalize the predicted and ground-truth translations ($\hat{\mathbf{T}}, \mathbf{T}_{\text{gt}}$) to purely penalize directional errors, setting $\epsilon = 10^{-6}$.

The final pose loss is the weighted sum over all layers:
\begin{align} \label{eq13}
\mathcal{L}_{\text{pose}} = \sum_{l=0}^{L-1} w_l \mathcal{L}^{\text{pose}}_l
\end{align}
where $L=3$ and the weights $w_l$ increase with depth to emphasize finer corrections from deeper layers $\mathbf{w} = [0.2, 0.3, 1.0]$.

\textbf{Uncertainty-aware Loss.} To further encourage meaningful dynamic masking without resorting to heuristic thresholding or auxiliary binary segmentation heads~\cite{teed2021droid,ye2023pvo,shen2023dytanvo}, we introduce an effective self-supervised uncertainty regularization. Instead of predicting per-pixel dynamic probabilities, we directly estimate a continuous uncertainty value $\mathbf{U}$ for each downsampled image patch (not per-pixel). This uncertainty is predicted from the concatenated query-patch features within the Motion-Semantic attention module.

We compute the photometric inconsistency induced by warping with the previous-frame flow estimate (same cosine distance used for the static confidence map $\mathbf{M}_t$):
\begin{align} \label{eq14}
\mathcal{L}_{l}^{u} = \frac{\min\left(\frac{1}{2}, 1 - \mathbf{M}_t\right)}{2\mathbf{U}_l^2} + \lambda \log \mathbf{U}_l, \qquad \mathcal{L}_u = \sum_{l=0}^{L-1} \mathcal{L}_{l}^{u}
\end{align}
where a similarity close to 1 indicates high confidence (likely static), while lower similarity suggests potential dynamic content, resulting in larger $U$. The normalized logarithmic term with $\lambda = 0.5$ ensures gradient stability.

The final visual odometry loss is a weighted combination:
\begin{align} \label{eq15}
\mathcal{L}_{\text{vo}} = \alpha \mathcal{L}_{\text{pose}} + \beta \mathcal{L}_u
\end{align}

\subsection{Training and Inference Pipeline}

\textbf{Training Phase.} As illustrated in \Cref{fig6}, we employ a two-stage training strategy. First, the Flow Backbone is pretrained independently using ground-truth optical flow to establish robust motion representations. Second, the full MVOFormer loads this pretrained backbone and jointly fine-tunes it alongside the MS Encoder. To minimize computational and memory overhead, optical flow is pre-computed and cached offline using a frozen Sea-RAFT model~\cite{wang2024sea}. Notably, the Warp Module is activated exclusively during training to provide self-supervisory signals for the Uncertainty Module.

\textbf{Inference Phase.} During deployment, MVOFormer directly regresses 6-DoF relative poses from raw image pairs in a single forward pass, entirely bypassing hand-crafted heuristics and classical backend optimization. This architectural simplicity is central to our model's efficiency and robustness.

\section{Experiments}
\label{sec4}
\subsection{Implementation Details}
\label{subsec4.1}
\textbf{Datasets and Evaluation.} To address the scarcity of dynamic scenes in the standard TartanAir dataset~\cite{wang2020tartanair}, we jointly train MVOFormer on both TartanAir and the dynamic Shibuya sequence~\cite{shen2023dytanvo}. We perform true zero-shot evaluation—without any target-domain fine-tuning—across KITTI~\cite{geiger2012we}, the TartanAir test split, TUM-RGBD~\cite{sturm2012benchmark}, and ETH3D-SLAM~\cite{schops2019bad}. Following~\cite{campos2021orb}, performance is measured by the Absolute Trajectory Error (ATE [m]) after $Sim(3)$  alignment of the estimated monocular trajectories to the ground truth.

\textbf{Configuration and Training.} Our network utilizes 8 attention heads, a 256-channel dimension, and 100 decoder queries. Implemented on an NVIDIA A100 GPU via the AdamW optimizer (weight decay $10^{-4}$), we adopt a two-stage training strategy (\Cref{fig6})~\cite{wang2021tartanvo}. First, the Flow Backbone is pretrained on TartanAir for 200 epochs (batch size 256, lr $4 \times 10^{-4}$, 0.1$\times$ decay at epochs 100 and 150). Second, the full model is fine-tuned for 50 epochs using offline-cached flow (batch size 64, lr $2 \times 10^{-4}$, 0.1$\times$ decay at epoch 30).

\begin{table*}[!t]
  \centering
  \caption{\centering \textsc{ATE $\downarrow$ Results on KITTI Odometry 0-10, Comparing with Other SLAM/VO Methods}}
  {
  \resizebox{\linewidth}{!}{\begin{threeparttable}
    \begin{tabular}{clccccccccccc|c}
    \toprule  
     \multicolumn{2}{c}{Methods} & 00    & 01    & 02    & 03    & 04    & 05    & 06    & 07    & 08    & 09    & 10    & Avg \\ \midrule 
    \multirow{4}[1]{*}{SLAM} & ORB-SLAM3~\cite{campos2021orb}‡ & 6.77  & \scalebox{0.6}\XSolidBrush     & 30.50 & 1.04  & 0.93  & 5.54  & 16.61 & 9.70  & 60.69 & 7.89  & 8.65  & \scalebox{0.6}\XSolidBrush \\
          & LDSO~\cite{gao2018ldso}‡  & 9.32  & 11.68 & 31.98 & 2.85  & 1.22  & 5.10  & 13.55 & 2.96  & 129.02 & 21.64 & 17.36 & 22.43 \\
          & DROID-SLAM~\cite{teed2021droid}‡ & 92.10 & 344.60 & \scalebox{0.6}\XSolidBrush     & 2.38  & 1.00  & 118.50 & 62.47 & 21.78 & 161.60 & \scalebox{0.6}\XSolidBrush & 118.70 & \scalebox{0.6}\XSolidBrush \\
          & DPV-SLAM++~\cite{lipson2024deep}‡ & 8.30  & 11.86 & 39.64 & 2.50  & 0.78  & 5.74  & 11.60 & 1.52  & 110.90 & 76.70 & 13.70 & 25.75 \\
    \midrule
    \multirow{9}[3]{*}{VO} & MambaVO~\cite{wang2025mambavo} & 112.39 & \textbf{8.16} & 93.78 & \textbf{1.80} & \textbf{0.66} & 56.51 & 57.19 & 17.90 & 116.01 & 73.56 & 14.37 & 50.21 \\
          & DROID-VO~\cite{teed2021droid} & 98.43 & 84.20 & 108.80 & 2.58  & 0.93  & 59.27 & 64.40 & 24.20 & 64.55 & 71.80 & 16.91 & 54.19 \\
          & DPVO~\cite{teed2023deep} & 113.21 & \uline{12.69} & 123.40 & \uline{2.09}  & \uline{0.68}  & 58.96 & 54.78 & 19.26 & 115.90 & 75.10 & 13.63 & 53.61 \\ 
          & DeepVO~\cite{wang2017deepvo} * & \scalebox{0.6}\XSolidBrush & 19.98 & \scalebox{0.6}\XSolidBrush & 11.74 & 3.85  & 123.30 & 108.00 & 22.83 & \scalebox{0.6}\XSolidBrush & \scalebox{0.6}\XSolidBrush & 57.90 & \scalebox{0.6}\XSolidBrush \\
          & TSformer-VO~\cite{zhao2022transformer} * & \scalebox{0.6}\XSolidBrush & 101.70 & \scalebox{0.6}\XSolidBrush  & 20.12 & 6.01  & 37.68 & 46.79 & 23.16 & \scalebox{0.6}\XSolidBrush & \scalebox{0.6}\XSolidBrush & 23.14 & \scalebox{0.6}\XSolidBrush \\
          & SWformer-VO~\cite{wu2024swformer} * & \scalebox{0.6}\XSolidBrush & 82.74 & \scalebox{0.6}\XSolidBrush & 14.83 & 4.37  & 50.15 & 24.26 & 28.12 & \scalebox{0.6}\XSolidBrush & \scalebox{0.6}\XSolidBrush & 16.97 & \scalebox{0.6}\XSolidBrush \\
           & TartanVO~\cite{wang2021tartanvo} † & 69.11 & 53.19 & \uline{78.78} & 2.70  & 1.99  & 55.18 & \textbf{10.50} & \uline{13.87} & 48.16 & 27.93 & 11.90 & 33.94 \\
          & DytanVO~\cite{shen2023dytanvo} † & \uline{41.79} & 13.06 & 83.97 & 2.17  & 1.33  & \uline{26.56} & 25.39 & 21.45 & \uline{38.43} & \textbf{11.89} & \uline{8.57} & \uline{24.96} \\
\cmidrule{2-14}          
        & MVOFormer (Ours) † & \textbf{22.37} & 53.70 & \textbf{67.15} & 3.13  & 1.63  & \textbf{9.91} & \uline{14.24} & \textbf{6.52} & \textbf{19.39} & \uline{13.84} & \textbf{3.81} & \textbf{19.61} \\ \bottomrule
    \end{tabular}
     \begin{tablenotes}
      \footnotesize
        \item ``*" Methods trained or fine-tuned on the KITTI dataset. ``†" learning-based Frame-to-Frame methods. Methods marked with (‡) use global optimization / loop closure. \textbf{Bold} indicates the best VO result, while \uline{underline} denotes the second-best result.
    \end{tablenotes}
  \end{threeparttable}}}
  \label{tab2}
\end{table*}

\subsection{Evaluated on Open-Source Benchmarks}
\label{subsec4.2}

\begin{figure}[!t]
\centering
\includegraphics[scale=0.37]{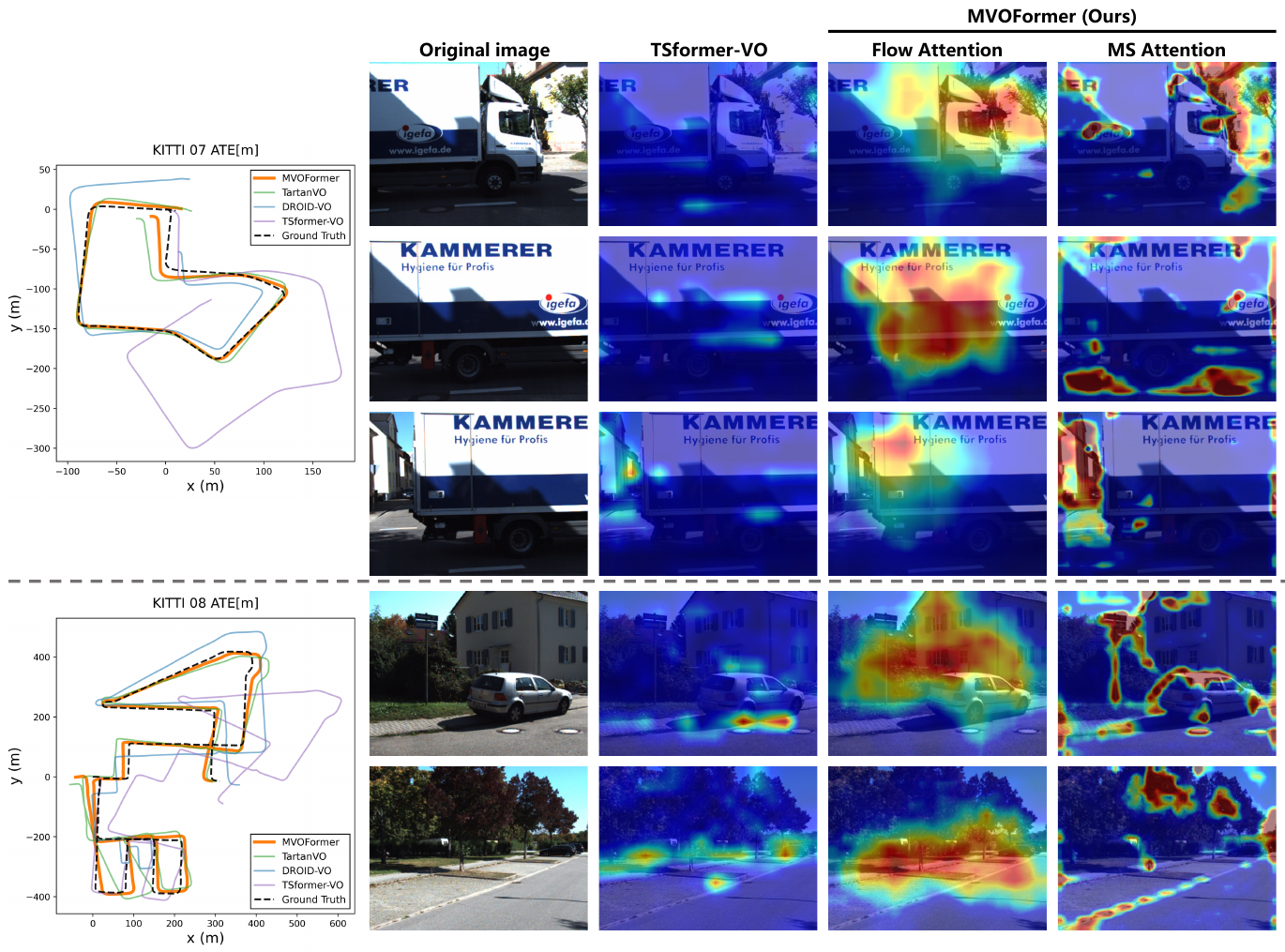}
\caption{Qualitative comparison of attention maps and motion trajectories on KITTI sequences.}
\label{fig7}
\end{figure}

\textbf{KITTI Dataset.} \Cref{tab2} reports the ATE on 11 real-world driving sequences. While traditional SLAM systems struggle in high-speed or feature-poor scenes, learning-based monocular VO demonstrates strong robustness. Notably, although it is evaluated strictly in a zero-shot setting (trained solely on TartanAir), MVOFormer outperforms all prior learning-based monocular VO approaches—including those explicitly fine-tuned on KITTI (marked with ``*"). It achieves the lowest average ATE, reducing it by 42.22\% compared with TartanVO.

\textbf{TartanAir Monocular Benchmark.} \Cref{tab1} evaluates 16 sequences featuring aggressive motion, extreme weather, and dynamic objects. While geometric and hybrid systems benefit from multi-frame backend optimization, we primarily compare against strictly learning-based Frame-to-Frame methods. Under such severe visual degradation, MVOFormer achieves the best average ATE among learning-based Frame-to-Frame methods and reduces it by 59.32\% compared with the previous leading method, TartanVO~\cite{wang2021tartanvo}.

\textbf{The TUM-RGBD and ETH3D-SLAM Datasets.} We further evaluate our method on indoor benchmarks using the same zero-shot model. We use sequences with ``fr1" camera intrinsics for TUM-RGBD and the ``1" training sequences for ETH3D-SLAM. As shown in \Cref{tab3}, MVOFormer reduces the average ATE by 27.75\% on TUM-RGBD and 24.14\% on ETH3D-SLAM compared with the previous best learning-based Frame-to-Frame method, while considerably narrowing the gap to multi-frame SLAM systems that exploit loop closure and bundle adjustment.

\textbf{Failure Cases.} MVOFormer exhibits reduced reliability on vegetation-heavy or highly cluttered sequences, such as KITTI 03 and 06, TartanAir MH005, and TUM-RGBD rpy. In these scenarios, dense foliage heavily degrades the underlying Sea-RAFT optical flow estimation, thereby destabilizing pose prediction. Furthermore, highly cluttered environments induce overly fragmented semantic representations. Constrained by the limited spatial resolution of the semantic tokens, the subsequent flow warping becomes imprecise, causing the uncertainty mask to erroneously over-suppress geometrically informative static regions. Future efforts will focus on joint flow-VO fine-tuning and efficiently scaling up token resolutions to enhance domain robustness.

\begin{table}[!t]
  \centering
  \caption{\centering \textsc{Performance Comparison of Various Methods on the TartanAir Monocular Test Split (ATE $\downarrow$).}}
  \resizebox{\columnwidth}{!}{\begin{threeparttable}
    \begin{tabular}{clc|cccccccc|c}
    \toprule  
    \multicolumn{2}{c}{Methods} & \makecell{ME\\Avg} & \makecell{ MH \\ 000}  & \makecell{ MH \\ 001}  & \makecell{ MH \\ 002}  & \makecell{ MH \\ 003}  & \makecell{ MH \\ 004}  & \makecell{ MH \\ 005}  & \makecell{ MH \\ 006}  & \makecell{ MH \\ 007}  & Avg \\
    \midrule
    \multicolumn{1}{c}{\multirow{5}[1]{*}{\rotatebox[origin=c]{90}{\scriptsize\makecell{MF\\(Geom./Hyb.)}}}} & ORB-SLAM3~\cite{campos2021orb}‡ & 15.99 & 15.44 & 2.92  & 13.51 & 8.18  & 2.59  & 21.91 & 11.70 & 25.88 & 14.38 \\
          & DSO~\cite{engel2017direct}   & 7.46 & 9.92  & 0.35  & 7.96  & 3.46  & \scalebox{0.6}\XSolidBrush & 12.58 & 8.42  & 7.50  & \scalebox{0.6}\XSolidBrush \\
          & COLMAP~\cite{schonberger2016structure}‡ & 9.80 & 12.26 & 13.45 & 13.45 & 20.95 & 24.97 & 16.79 & 7.01  & 7.97  & 12.21 \\
          & DROID-VO~\cite{teed2021droid} & 0.64 & 0.32  & 0.13  & 0.08  & 0.09  & 1.52  & 0.69  & 0.39  & 0.97  & 0.58 \\
          & DPVO~\cite{teed2023deep} & 0.24 & 0.21  & 0.04  & 0.04  & 0.08  & 0.58  & 0.17  & 0.11  & 0.15  & 0.21 \\
    \midrule
    \multicolumn{1}{c}{\multirow{4}[1]{*}{\rotatebox[origin=c]{90}{\scriptsize\makecell{F2F\\(Learning)}}}} 
    & DiffPoseNet~\cite{parameshwara2022diffposenet} & \scalebox{0.6}\XSolidBrush & 2.56  & 0.31  & 1.57  & \uline{0.72} & \uline{0.82}  & 1.83  & \uline{1.32}  & \uline{1.24}  & \scalebox{0.6}\XSolidBrush \\  
    & TartanVO~\cite{wang2021tartanvo} & \uline{5.39} & \uline{2.12} & 0.31  & \uline{1.28}  & 1.09  & 0.99  & \uline{1.40}  & 1.74  & 1.42  & \uline{3.34} \\
          & DytanVO~\cite{shen2023dytanvo} & 5.56 & 5.10  & \textbf{0.22} & 1.62  & 0.79  & 1.29  & 4.46  & 2.06  & 2.36  & 3.90 \\
          \cmidrule{2-12}
          & MVOFormer (Ours) & \textbf{1.81} & \textbf{0.91}  & \uline{0.25}  & \textbf{0.89} & \textbf{0.58}  & \textbf{0.40} & \textbf{1.37} & \textbf{1.20} & \textbf{1.01} & \textbf{1.36} \\ \bottomrule
    \end{tabular}
     \begin{tablenotes}
      \footnotesize
       \item MF (Geom./Hyb.) denotes Multi-Frame (Geometric/Hybrid), and F2F (Learning) denotes learning-based Frame-to-Frame methods. \textbf{Bold} indicates the best F2F result, while \uline{underline} indicates the second-best result.
    \end{tablenotes}
  \end{threeparttable}}
  \label{tab1}
\end{table}

\begin{table*}[htbp]
  \centering
  \caption{\centering \textsc{Performance Comparison of Various Methods on the TUM-RGBD and ETH3D-SLAM (ATE $\downarrow$).}}
  \resizebox{\linewidth}{!}{\begin{threeparttable}
    \begin{tabular}{clccccccccc|c|c}
    \toprule  
    \multicolumn{2}{c}{\multirow{2}[2]{*}{Methods}} & \multicolumn{10}{c}{TUM-RGBD}  & \multicolumn{1}{l}{ETH3D-SLAM} \\
\cmidrule{3-13}    \multicolumn{2}{c}{}  & 360   & desk & desk2 & floor & plant & room & rpy & teddy & xyz & Avg  & Avg \\ \midrule
    \multicolumn{1}{c}{\multirow{4}[1]{*}{\makecell{ MF \\ (Geom./Hyb.)}}} & ORB-SLAM3~\cite{campos2021orb}‡ & \scalebox{0.6}\XSolidBrush & 0.016 & \scalebox{0.6}\XSolidBrush & \scalebox{0.6}\XSolidBrush & 0.038 & \scalebox{0.6}\XSolidBrush & \scalebox{0.6}\XSolidBrush & 0.145 & 0.005 & \scalebox{0.6}\XSolidBrush  & \scalebox{0.6}\XSolidBrush \\
          & DSO~\cite{engel2017direct} & \scalebox{0.6}\XSolidBrush & 0.405 & 0.322 & 0.041 & 0.108 & 0.800 & \scalebox{0.6}\XSolidBrush & \scalebox{0.6}\XSolidBrush & 0.058 & \scalebox{0.6}\XSolidBrush & \scalebox{0.6}\XSolidBrush \\
          & DROID-VO~\cite{teed2021droid} & 0.141 & 0.064 & 0.078 & 0.063 & 0.041 & 0.393 & 0.030 & 0.221 & 0.017 & 0.116 & 0.238 \\
          & DPVO~\cite{teed2023deep} & 0.156 & 0.034 & 0.050 & 0.183 & 0.034 & 0.383 & 0.038 & 0.073 & 0.012 & 0.107 & 0.203 \\
    \midrule
    \multicolumn{1}{c}{\multirow{4}[1]{*}{\makecell{ F2F \\ (Learning)}}} 
          & TrianFlow~\cite{zhao2020towards} & 0.187 & 0.526 & 0.483 & 0.739 & 0.388 & 0.884 & \textbf{0.050} & 0.554 & 0.182 & 0.444 & 0.706 \\
          & TartanVO~\cite{wang2021tartanvo} & \uline{0.160} & 0.478 & 0.539 & 0.348 & 0.395 & \uline{0.417} & \textbf{0.050} & \uline{0.329} & 0.160 & 0.320 & 0.421 \\
          & DytanVO~\cite{shen2023dytanvo} & 0.188 & \uline{0.159} & \uline{0.224} & \textbf{0.191} & \uline{0.343} & 0.530 & \uline{0.053} & 0.508 & \uline{0.131} & \uline{0.259} & \uline{0.364} \\ \cmidrule{2-13} 
          & MVOFormer (Ours) & \textbf{0.130} & \textbf{0.095} & \textbf{0.122} & \uline{0.286} & \textbf{0.302} & \textbf{0.294} & 0.089 & \textbf{0.297} & \textbf{0.069} & \textbf{0.187} & \textbf{0.276} \\ \bottomrule
    \end{tabular}
    \end{threeparttable}}
  \label{tab3}
\end{table*}

\begin{table}[!t]
  \centering
   \caption{\centering \textsc{Comparison of Inference Efficiency (Parameters, Memory, and FPS)}}
    \resizebox{\linewidth}{!}{\begin{threeparttable}
    \begin{tabular}{lcccc}
    \toprule  
        Methods     & Parameters (M) & GPU Memory (GB) & FPS   &  \\ \midrule
        DROID-VO    & 4.00       & 21.22           & 6.36  &  \\
        TartanVO    & 24.51      & 0.22            & 80.03 &  \\
        DytanVO     & 163.04     & 3.20            & 6.18  &  \\
        TSformer-VO & 30.66      & 0.54            & 15.07 &  \\  \midrule
        MVOFormer (Ours)    & 90.86      & 1.02     & 32.19 &  \\ \bottomrule
    \end{tabular}
    \end{threeparttable}}
  \label{tab6}
\end{table}

\begin{table}[!t]
  \centering
  \caption{\centering \textsc{Module Ablation on MVOFormer.}}
    \resizebox{\linewidth}{!}{\begin{threeparttable}
    \begin{tabular}{cccccc}
    \toprule  
    \multicolumn{2}{c}{Configuration} & TartanAir & KITTI & TUM-RGBD & Avg \\
    \midrule
    \multicolumn{1}{c}{\makecell{Baseline}} & \multicolumn{1}{l}{TartanVO~\cite{wang2021tartanvo}} & 3.34 & 33.94 & 0.320 & 12.533 \\
    \midrule
    \multicolumn{1}{c}{\multirow{3}[0]{*}{\makecell{ Flow Backbone \\ (w/o Semantic Branch)}}} & \multicolumn{1}{l}{$N = 50$}  & 2.696 & 32.175 & 0.254 & 11.708 \\
          & \multicolumn{1}{l}{$N = 100$} & 2.667 & 30.595 & 0.259 & 11.174 \\
          & \multicolumn{1}{l}{$N = 500$} & 2.420 & 41.571 & 0.266 & 14.752 \\
    \midrule
    \multirow{4}[1]{*}{\makecell{ MVOFormer \\ (w. Semantic Branch)}} 
          & \multicolumn{1}{l}{w/o IP, w/o UM} & 1.980 & 28.020 & 0.254 & 10.085 \\
          & \multicolumn{1}{l}{w. IP only} & 1.979 & 25.759 & 0.232 & 9.323 \\
          & \multicolumn{1}{l}{w. UM only} & \uline{1.516} & \uline{20.109} & \uline{0.191} & \uline{7.272} \\
           \cmidrule{2-6}
          & \multicolumn{1}{l}{Full model} & \textbf{1.359} & \textbf{19.609} & \textbf{0.187} & \textbf{7.052} \\
    \bottomrule
    \end{tabular}

         \begin{tablenotes}
      \footnotesize
        \item UM: Uncertainty Module; IP: Iterative Pose Refinement.
    \end{tablenotes}
  \end{threeparttable}}
  
  \label{tab5}
\end{table}

\begin{table}[!t]
  \centering
  \caption{\centering \textsc{Loss Weight Ablation.}}
    \resizebox{\linewidth}{!}{\begin{threeparttable}
    \begin{tabular}{llccccc}
    \toprule
    Model & $\lambda$ & $(\alpha, \beta)$ & TartanAir & KITTI & TUM-RGBD & Avg \\
    \midrule
    \multirow{7}[1]{*}{\hspace{5pt}\rotatebox[origin=c]{90}{MVOFormer}} & \multirow{3}[0]{*}{0.1} & $(0.5, 0.1)$ & 1.231 & 31.272 & 0.267 & 10.923 \\
          &  & $(0.5, 0.3)$ & 1.225 & 30.408 & \uline{0.236} & 10.623 \\
          &  & $(0.5, 0.5)$ & 1.342 & 32.971 & 0.241 & 11.518 \\
    \cmidrule{2-7}
          & 0.3 & $(0.5, 0.3)$ & \uline{1.209} & 29.995 & 0.249 & 10.484 \\
          & 0.5 & $(0.5, 0.3)$ & \textbf{1.173} & \uline{28.885} & 0.209 & \uline{10.089} \\
          & 1.0 & $(0.5, 0.3)$ & 1.583 & 30.288 & \uline{0.202} & 10.691 \\
          & \cellcolor{gray!15}0.8 & \cellcolor{gray!15}$(0.5, 0.3)$ & \cellcolor{gray!15}{1.359} & \cellcolor{gray!15}\textbf{19.609} & \cellcolor{gray!15}\textbf{0.187} & \cellcolor{gray!15}\textbf{7.052} \\
    \bottomrule
    \end{tabular}

         \begin{tablenotes}
      \footnotesize
        \item $\lambda$ is the uncertainty regularization coefficient in \eqref{eq14}, while $\alpha$ and $\beta$ correspond to the loss weights in \eqref{eq15}. The last gray row denotes final MVOFormer.
    \end{tablenotes}
  \end{threeparttable}}

  \label{tab5b}
\end{table}

\textbf{Qualitative Analysis.} \Cref{fig7} visualizes attention maps (brighter regions indicate higher weights) from the final decoder block on KITTI. Notably, MVOFormer is evaluated strictly zero-shot, whereas the baseline TSformer-VO was trained in-domain. Compared to TSformer-VO's space-time attention, our Flow Attention already reduces distractions from moving vehicles. Furthermore, the fused Motion-Semantic (MS) Attention largely ignores dynamic objects and focuses on geometrically stable and semantically meaningful static structures (e.g., roads and lane markings), even under partial foliage occlusion. This behavior highlights the object-level understanding provided by DINOv3. Finally, the plotted x-y motion trajectories confirm MVOFormer's strong resistance to drift and superior generalization on unseen datasets.

\subsection{Efficiency Analysis}

To ensure a rigorous comparison, we standardize our efficiency evaluation on an NVIDIA RTX 4090 GPU and an Intel i7-14700KF CPU with a batch size of 1. \Cref{tab6} details the computational efficiency. For MVOFormer, the standard image input is $480 \times 640$ (TartanAir native; KITTI center-cropped). Internally, the optical flow is downsampled to $120 \times 160$ before entering the encoder, and DINOv3 inputs are resized to $518 \times 518$. To ensure a fair frame-to-frame comparison, baseline models strictly adopt their default framework settings, with DROID-VO explicitly evaluated without global bundle adjustment. Under these configurations, MVOFormer achieves 32.19 FPS (averaged across test splits), exceeding the 10 Hz real-time requirement of KITTI benchmarks.

\subsection{Ablation Studies}
\label{subsec4.4}

We summarize the ablation results on TartanAir, KITTI, and TUM-RGBD in \Cref{tab5,tab5b}. For the flow backbone, increasing the number of queries from 50 to 100 consistently improves performance. However, a larger setting ($N=500$) causes overfitting and degrades generalization. We therefore adopt $N=100$. As shown in \Cref{tab5}, adding the semantic branch improves all three benchmarks over the pure flow baseline. On this backbone, IP is most beneficial on KITTI, while UM is more effective on TartanAir. Combining IP and UM gives the best overall trade-off.

For the loss weights in \eqref{eq14} and \eqref{eq15}, \Cref{tab5b} shows that $(\alpha, \beta) = (0.5, 0.3)$ performs best. An excessively large $\beta$ over-emphasizes uncertainty and weakens pose estimation. In contrast, a smaller $\beta$ makes the uncertainty effect insufficient. The regularization weight $\lambda=0.8$ also gives the best trade-off. A higher $\lambda$ makes uncertainty dominate optimization, whereas a lower $\lambda$ cannot adequately suppress dynamic regions.

\begin{figure}[!t]
\centering
\includegraphics[width=\linewidth]{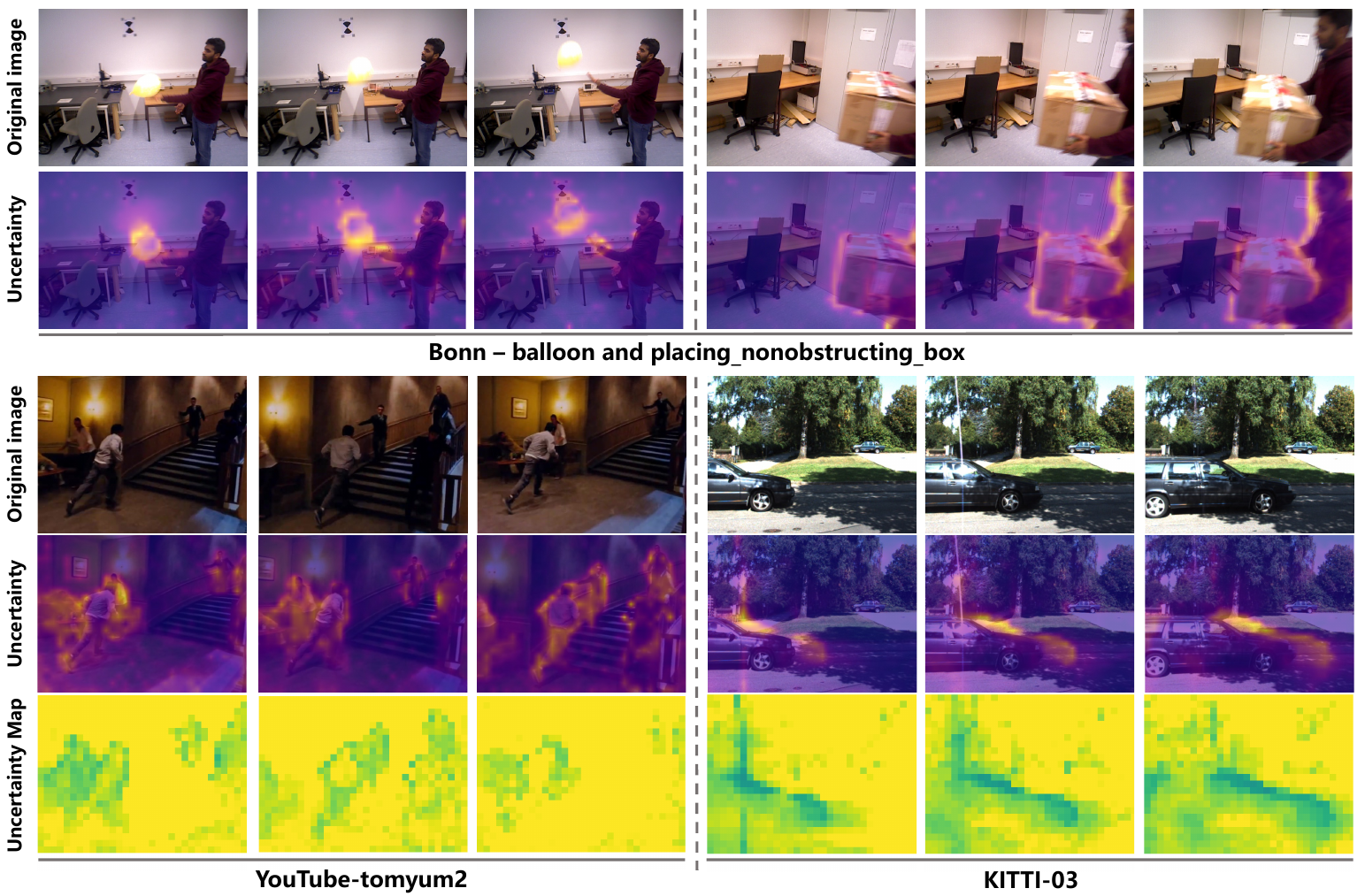}
\caption{Uncertainty evaluation on dynamic sequences.}
\label{fig10_2}
\end{figure}

\begin{table}[!t]
  \centering
  \caption{\centering \textsc{Ablation on the Dynamic Bonn Dataset (ATE $\downarrow$).}}
    \resizebox{\linewidth}{!}{\begin{threeparttable}
    \begin{tabular}{lcccccccc}
    \toprule
    Configuration & balloon & BT & PT & PNO & PO & RNO & RO & Avg \\
    \midrule
    w/o UM, w/o IP & 0.127 & 0.195 & 0.162 & 0.121 & 0.175 & 0.085 & 0.123 & 0.141 \\
    w. IP only & 0.125 & 0.169 & 0.163 & 0.116 & 0.167 & 0.069 & 0.114 & 0.132 \\
    w. UM only & \uline{0.121} & \uline{0.164} & \uline{0.135} & \uline{0.102} & \uline{0.137} & \uline{0.065} & \uline{0.080} & \uline{0.115} \\
    \rowcolor{gray!15}
    Full model & \textbf{0.117} & \textbf{0.110} & \textbf{0.130} & \textbf{0.079} & \textbf{0.128} & \textbf{0.062} & \textbf{0.065} & \textbf{0.099} \\
    \bottomrule
    \end{tabular}
         \begin{tablenotes}
      \footnotesize
        \item BT/PT denote balloon tracking/person tracking; PNO/PO: placing (nonobstructing)/(obstructing) box; RNO/RO: removing (nonobstructing)/(obstructing) box. The shaded final row corresponds to the Full model with $\lambda=0.5$.
    \end{tablenotes}
    \end{threeparttable}}
  \label{tab6b}
\end{table}

Dynamic-scene ablations on the Bonn RGB-D dynamic dataset are reported separately in \Cref{tab6b}. Introducing the Uncertainty Module alone reduces the error by 18.44\% relative to the baseline. Adding iterative pose refinement further increases the improvement to 29.79\%. As shown in \Cref{fig10_2}, the warp-supervised UM highlights subtle motion boundaries even under small inter-frame displacement. The estimated uncertainty also increases with motion magnitude over larger temporal intervals.

\section{Conclusion}
\label{sec5}

\textbf{Limitations and Future Work.} As a strictly frame-to-frame monocular system, MVOFormer remains vulnerable to accumulated drift compared with global SLAM backends. Future work will extend the framework to stereo and RGB-D modalities through either joint prediction models analogous to FoundationStereo~\cite{wen2025foundationstereo} or explicit depth-prior fusion following TartanVO-Stereo~\cite{qiu2025mac}. In addition, integrating a lightweight multi-frame optimization module may further mitigate long-term trajectory drift.

\textbf{Conclusion.} In this paper, we introduced MVOFormer, a transformer-based framework for robust monocular visual odometry. By combining pixel-level optical flow with object-aware motion-semantic priors within a unified architecture, MVOFormer achieves iterative pose refinement and robust dynamic distractor suppression while operating at 32.19 FPS. Extensive evaluations across TartanAir, KITTI, TUM-RGBD, and ETH3D-SLAM confirm that our method consistently surpasses existing learning-based frame-to-frame monocular VO baselines under a strict zero-shot protocol, requiring neither dataset-specific fine-tuning nor classical post-processing. We expect MVOFormer to provide a robust and reproducible foundation for future research in learning-based visual odometry.

\balance

\end{document}